\documentclass[10pt,twocolumn,letterpaper]{article}

\usepackage[pagenumbers]{cvpr} 
\usepackage{multirow}
\definecolor{cvprblue}{rgb}{0.21,0.49,0.74}
\usepackage[pagebackref,breaklinks,colorlinks,allcolors=cvprblue]{hyperref}

\newcommand{\wlink}[1]{\textcolor{magenta}{{#1}}}

\title{MCMat: Multiview-Consistent and Physically Accurate \\ 
PBR Material Generation}

\author{Shenhao Zhu$^{1,2}\footnotemark[1]$ \quad Lingteng Qiu$^{2}\footnotemark[1]$ \quad Xiaodong Gu$^{2}\footnotemark[1]$ \quad Zhengyi Zhao$^{2}\footnotemark[1]$ \\ \quad Chao Xu$^{2}$  \quad Yuxiao He$^{1}$ \quad Zhe Li$^{2,3}$ \quad Xiaoguang Han$^{5,6}$ \quad Yao Yao$^{1}$ \quad Xun Cao$^{1}$ \\ \quad Siyu Zhu$^{4}$ \quad Weihao Yuan$^{2}$ \quad Zilong Dong$^{2}\footnotemark[2]$ \quad Hao Zhu$^{1}\footnotemark[2]$
\vspace{0.3em} \\
{\normalsize $^1$Nanjing University}
\quad{\normalsize $^2$Alibaba Group}  \quad {\normalsize $^3$Huazhong University of Science and Technology} \\ \quad{\normalsize $^4$Fudan University} \quad{\normalsize $^5$SSE, CUHKSZ} \quad{\normalsize $^6$FNii, CUHKSZ}
}

\begin{document}
\twocolumn[{
\renewcommand\twocolumn[1][]{#1}
\maketitle
\vspace{-18pt}
\begin{center}
    \captionsetup{type=figure}
    \includegraphics[width=\textwidth]{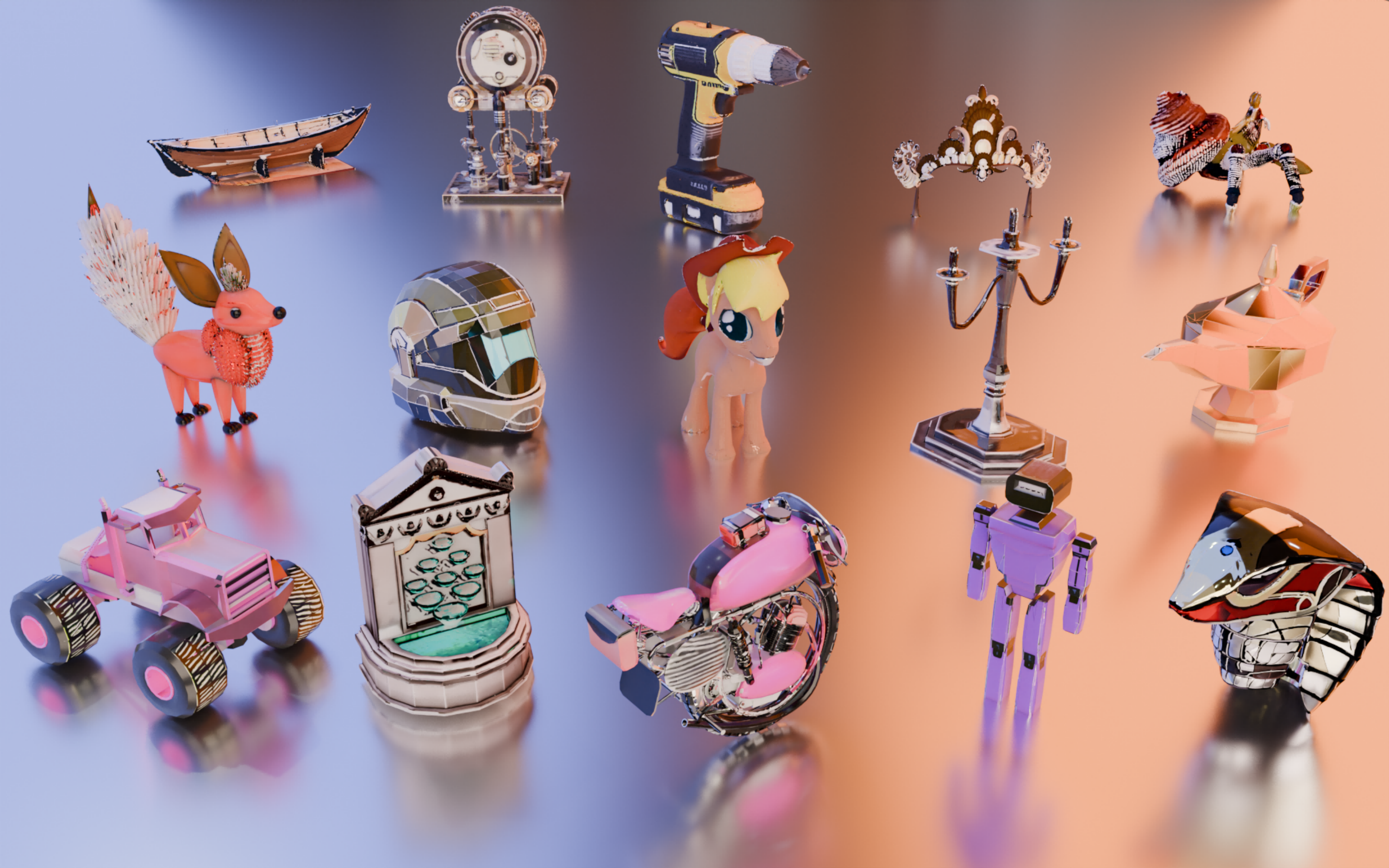}
    \captionof{figure}{\textbf{A gallery of generated textured meshes.} Our method effectively produces high-quality, lighting-independent, and highly faithful PBR materials across a wide range of objects from various categories, achieving highly realistic rendering results.}
    \label{fig:teaser}
\end{center}
}]
\footnotetext[1]{Equal contribution.}
\footnotetext[2]{Corresponding author.}


\begin{abstract}
Existing 2D methods utilize UNet-based diffusion models to generate multi-view physically-based rendering (PBR) maps but struggle with multi-view inconsistency, while some 3D methods directly generate UV maps, encountering generalization issues due to the limited 3D data.
To address these problems, we propose a two-stage approach, including multi-view generation and UV materials refinement.
In the generation stage, we adopt a Diffusion Transformer (DiT) model to generate PBR materials, where both the specially designed multi-branch DiT and reference-based DiT blocks adopt a global attention mechanism to promote feature interaction and fusion between different views, thereby improving multi-view consistency. 
In addition, we adopt a PBR-based diffusion loss to ensure that the generated materials align with realistic physical principles. 
In the refinement stage, we propose a material-refined DiT that performs inpainting in empty areas and enhances details in UV space. 
Except for the normal condition, this refinement also takes the material map from the generation stage as an additional condition to reduce the learning difficulty and improve generalization. 
Extensive experiments show that our method achieves state-of-the-art performance in texturing 3D objects with PBR materials and provides significant advantages for graphics relighting applications. Project Page: \wlink{https://lingtengqiu.github.io/2024/MCMat/}

\end{abstract}

\section{Introduction}
Generating physically-based rendering (PBR) materials for a given 3D model according to a given text prompt or an image can greatly reduce the cost of manual creation, which offers significant value across games, movies, and VR/AR. Compared to the RGB texture, 3D models with the PBR materials can be relighted in new lighting environments following graphical rendering pipelines to generate more realistic effects.

To achieve broad generalization in generating texture maps from arbitrary text prompts, recent studies have leveraged pre-trained 2D diffusion models. These works~\cite{zero123, MVDiffusion_Tang_Zhang_Chen_Wang_Furukawa} employ the 2D diffusion model to generate high-quality images corresponding to the given texts and additional depth conditions.
However, the pre-trained 2D diffusion models only generate 2D results in the current view rather than the completed texture map. Some approaches\cite{TEXTure_10.1145/3588432.3591503, chen2023text2tex} have attempted to address this limitation through a progressive inpainting strategy. These methods begin by generating a reference image from a predefined viewpoint, which is then projected back to the UV space to create a partial texture map. The process is iteratively repeated from new viewpoints until all predefined views are covered. Unfortunately, such methods can suffer from cumulative inconsistencies, leading to potential artifacts in the final texture map. Subsequent research~\cite{synvmvd_DBLP:journals/corr/abs-2311-12891} has aimed to tackle this issue by fine-tuning 2D diffusion models for simultaneous multi-view generation. However, these multiview-based methods only generate illumination-coupled texture maps, which can adversely affect rendering in new lighting environments. 

In this paper, we propose a multi-view PBR materials generator based on the video diffusion transformer model (DiT)~\cite{Peebles_2023,hong2022cogvideo_t2v, yang2024cogvideoxtexttovideodiffusionmodels} which has a billion parameters trained on large-scale video data to ensure robust generalization.
This generator incorporates a cross-frame global attention mechanism to facilitate feature interaction and fusion across different views, improving spatial consistency in generated outputs. 
Specifically, we introduce a multi-branch DiT fine-tuned from the video DiT model, which utilizes geometric guidance, i.e., surface normal information from the 3D model as geometric control conditions to generate albedo, roughness, and metallic material properties. To improve the accuracy of PBR material generation, we incorporate a PBR-based diffusion loss, ensuring that the generated materials are consistent with realistic physical principles. Additionally, to enhance the preservation of detailed reference information, we design a reference-based DiT block that applies appearance self-attention to fuse the reference information into each view.

After generating multi-view outputs and back-projecting them to create initial UV material map, two primary issues arise. First, the multi-view results fail to completely cover the mesh, resulting in empty areas within the coarse material map. Second, the relatively low-resolution multi-view images do not provide the necessary details for a high-resolution material map, leading to a loss of fine details.
To address these two challenges, we propose a material-refined DiT that refines the material maps further. This DiT uses the coarse material map as a condition, performing inpainting in the void regions while enhancing details in UV space. This process ultimately results in high-quality 2K resolution material images.
Notably, this refined learning process is simpler compared to some recent works~\cite{Yu_2023_ICCV_point_uv, guv_Texturify} that directly predict high-fidelity texture maps without any coarse initialization. 
We summarize our main contributions as follows:
\begin{itemize}
    \item We propose a Multi-View Generation DiT (MG-DiT) for generating PBR materials. Our approach incorporates a specially designed multi-branch DiT and a reference-based DiT block, enabling the generation of multiple consistent multi-view materials while enhancing the preservation of detailed reference information. Additionally, we employ a PBR-based diffusion loss to ensure that the generated materials align with realistic physical principles.
    \item We propose a Material Refinement DiT (MR-DiT) that not only performs inpainting in empty regions but also enhances details in UV space. This culminates in the production of high-quality materials at 2K resolution.
    \item Extensive experiments demonstrate that our method achieves state-of-the-art performance in the texturing of 3D objects with PBR materials conditioned on various text or image inputs and offers significant advantages for graphics relighting applications.
\end{itemize}

\section{Related Work}

\paragraph{Multi-View Based Texturing} 
2D UNet-based Diffusion models~\cite{sd,control-net} have achieved significant success in Text-to-Image tasks.
Some methods~\cite{MVDiffusion_Tang_Zhang_Chen_Wang_Furukawa,latent-nerf,synvmvd_DBLP:journals/corr/abs-2311-12891,zhao2024optimization_consisntex,richdreamer,MVDREAM_Shi_Wang_Ye_Mai_Li_Yang_Bytedance} utilize it to synthesize texture for 3D objects. 

TEXTure~\cite{TEXTure_10.1145/3588432.3591503} has designed an incremental pipeline based on view inpainting. Depth-maps of all viewpoints are rendered from the mesh to serve as the condition of the T2I pretrained model. Similarly, Text2Tex ~\cite{chen2023text2tex} generates texture iteratively and develops an automatic selection method to refine the final UV-map. Although the textures on the front side of the models are often high-quality, the results on the back side are not satisfactory due to the lack of global information modeling in these pipeline.  TexFusion~\cite{TexFusion_Cao_2023_ICCV} proposes aggregating information from different viewpoints during denoising to synthesize an entire latent-uv map, improving the consistency of viewing results. Later, several methods~\cite{synvmvd_DBLP:journals/corr/abs-2311-12891,zero123pp} replace serial pipelines with parallel ones to generate more consistent textures. However, results generated by these methods are limited to albedo maps, lacking more comprehensive attributes, which are difficult to apply in the industrial production processes. 

TANGO~\cite{TANGO} uses a Spherical Gaussian-based differentiable renderer and synthesizes materials via CLIP loss. Fantasia3D~\cite{Chen_2023_ICCV_Fantasia3D} disentangles modeling and learning of geometry and appearance. Paint-it~\cite{youwang2023paintit} generates PBR materials via a neural re-parameterized texture optimization designed on SDS loss~\cite{dreamfusion}. FlashTex~\cite{deng2024flashtex} introduces a T2I model conditioned on a rendered lighting map on ControlNet~\cite{control-net} architecture, which promotes the generation of PBR material. However, these works still struggle to generate high-quality material on complex cases due to the local attention in UNet-based diffusion models. We adopt the transformer-based diffusion model with global attention to improve consistency and visual quality.


\paragraph{3D Texture Generation}

Some pioneering methods~\cite{Objaverse,Jun_Nichol_shapee,pointe_Nichol_Jun_Dhariwal_Mishkin_Chen_2022,CoMPaT_Li_Upadhyay_Slim_Abdelreheem_Prajapati_Pothigara_Wonka_Elhoseiny_2022,luo2023scalable,GPLD3D_Dong_2024_CVPR,Texture_Fields_Oechsle_Mescheder_Niemeyer_Strauss_Geiger_2019,GET3D_Gao_Shen_Wang_Chen_Yin_Li_Litany_Gojcic_Fidler,3DGen_Gupta_Xiong_Nie} assign colors to surface points of 3D models to generate textures instead of fully exploiting the topological relationships of the surface mesh. 
Subsequently, several methods~\cite{guv_Bokhovkin_Tulsiani_Dai_2023,guv_Chen_Yin_Fidler_2022,guv_Cheng_Li_Liu_Wang,guv_Karras_Aittala_Laine_Härkönen_Hellsten_Lehtinen_Aila_2021,guv_Texturify,guv_Yu_Dong, he2024head360} have specifically designed and characterized spaces for texture generation, with the use of UV maps as carriers for texture generation gradually becoming a consensus. Approaches like PointUV~\cite{Yu_2023_ICCV_point_uv} and Paint3D~\cite{zeng2023paint3d} directly perform inpainting and color correction on the UV maps via training a diffusion model from the 3D dataset. These methods are hindered by the lack of large-scale high-quality 3D training datasets. Our approach provides a better coarse initialization to ensure the quality of the textures.


\paragraph{Video Diffusion Models} 


The rapid development of text-to-video models is propelled by advancements in Transformer architectures~\cite{vaswani2017attention_transformer} and diffusion models~\cite{ho2020denoising_diffusion}. Early efforts~\cite{ho2022imagen_t2v,hong2022cogvideo_t2v,villegas2022phenaki_t2v,singer2022makeavideo_t2v} demonstrated promising results, showcasing the potential of these approaches for generating realistic and coherent video sequences.
DiT architecture~\cite{Peebles_2023} has emerged as a powerful framework for video generation~\cite{SORA_videoworldsimulators2024}. Its global attention mechanism and robust generalization capabilities have pushed the boundaries of video synthesis, producing temporally consistent and visually appealing results.
Building on these advancements, recent works~\cite{zuo2024videomv,voleti2024sv3d} have extended video generation models into the 3D domain. These methods leverage the strengths of video diffusion models to generate high-quality 3D models with detailed textures and realistic material properties, bridging the gap between video and 3D synthesis.
Inspired by this progress, we introduce MG-DiT to enable PBR material generation. MG-DiT capitalizes on the global attention and generalization capabilities of video diffusion models, ensuring multi-view consistency and enhancing the quality of generated PBR materials.


\section{Method}

\begin{figure*}[th]
    \centering
    \includegraphics[width=1.0\linewidth]{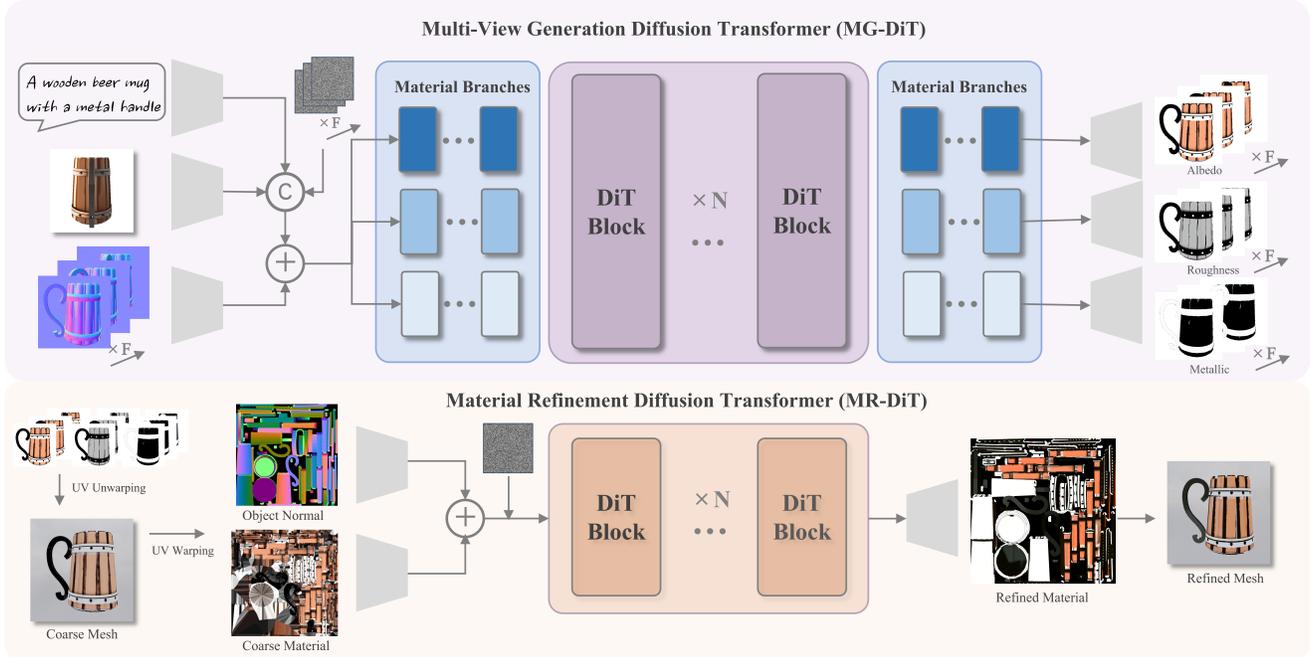}
    \caption{Our method consists of a generation stage and a refinement stage. In the generation stage, the Multi-View Generation DiT (MG-DiT) model utilizes surface normal information from the 3D model as geometric conditions, reference images, and textual descriptions to generate multi-view-consistent PBR material properties. In the refinement stage, the Material Refinement DiT (MR-DiT) model performs inpainting in void regions and enhances details in UV space, ultimately producing high-quality 2K resolution textures with precise material information.}
    \label{fig:pipeline}
\end{figure*}

Given a reference image or a piece of text and a corresponding 3D model represented as a triangular mesh, our objective is to transform this 3D model into a textured mesh with physically-based rendering (PBR) material properties based on details extracted from the given reference image or textual description.

Our method divides this task into a generation stage and a refining stage. In the generation stage, we propose Multi-View Generation DiT (MG-DiT), a model specifically designed for maintaining multi-view consistency in PBR material generation (Sec. \ref{sec:mvdit}), which utilizes the surface normal information from the 3D model as geometric conditions to generate multi-view-consistent PBR material properties for the object based on reference images and textual descriptions. Meanwhile, we employ a PBR-based Diffusion Loss (Sec. \ref{sec:pbrloss}) to further improve PBR material generation accuracy. In the refining stage, a Material Refinement DiT (MR-DiT) model is used to perform inpainting in void regions and further enhance details in UV space (Sec. \ref{sec:uvdit}), ultimately producing high-quality 2K resolution texture images with precise material information.

\subsection{Multi-View Generation DiT}
\label{sec:mvdit}
Our MG-DiT generates multi-view results of objects based on a single reference image and its corresponding textual description as appearance reference information. A 3D mesh $\mathcal{M}$ is provided as a known condition in the PBR material generation task, introducing geometric constraints to the generated results. Thus, MG-DiT incorporates appearance reference and geometric constraint information to generate diverse PBR materials through multiple noisy latents and a branched DiT structure. Additionally, it innovatively employs a reference-based DiT block to enhance 3D consistency while preserving reference information.

\begin{figure}
    \includegraphics[width=0.999\linewidth]{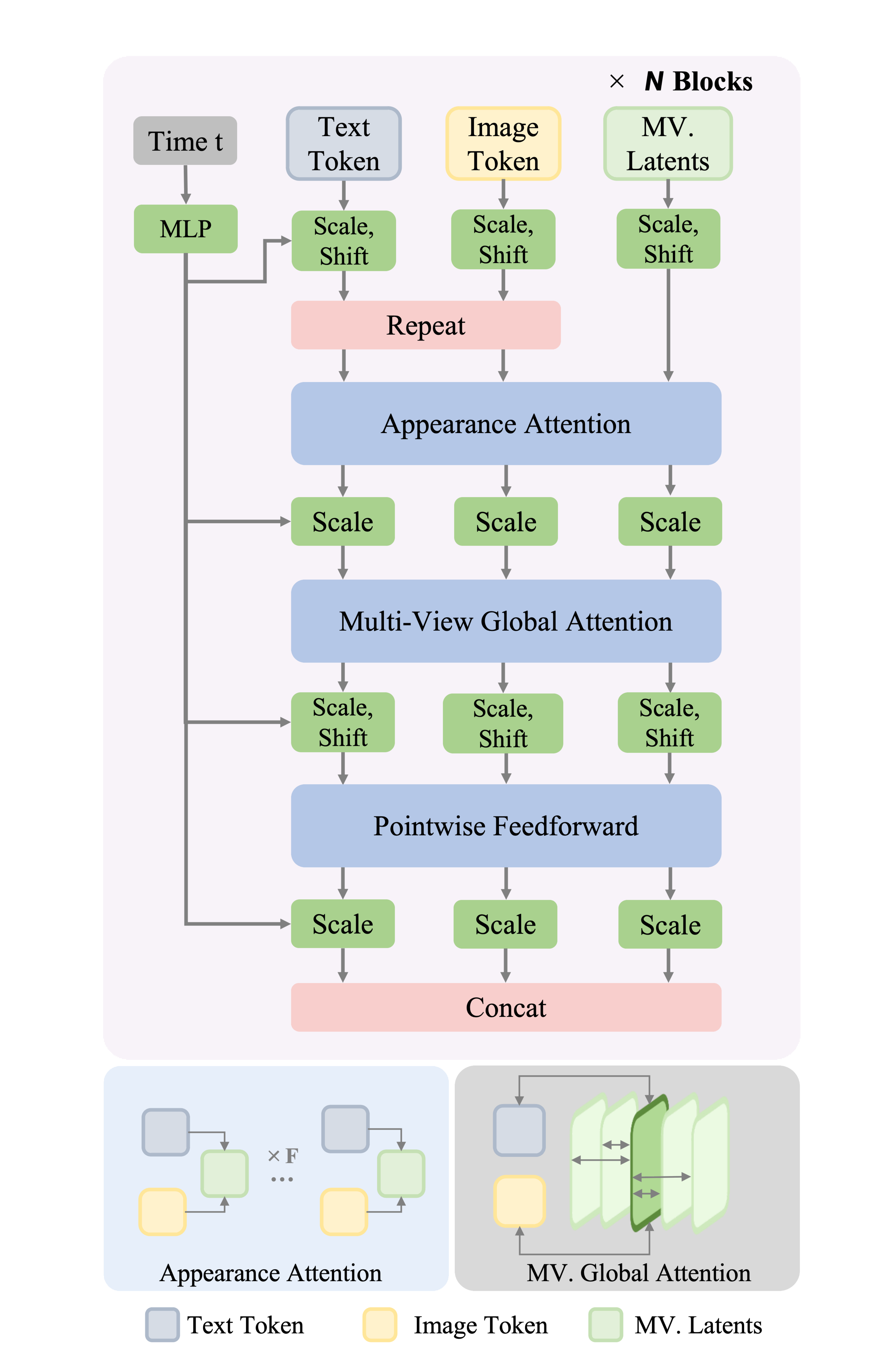}
    \caption{Structure of our Multi-View Generation DiT block.} \label{fig:block}
\end{figure}

\paragraph{Reference-based DiT Block.} Our material generation block adopts the structure shown in Fig. \ref{fig:pipeline}. The DiT generates PBR materials using both text and image modalities as input. When only text condition is provided, we leverage ControlNet~\cite{control-net} to generate the corresponding reference image with geometric details. Similarly, if only a reference image is given, Caption3D~\cite{luo2023scalable,luo2024view} produces the corresponding text prompt. The text prompt and reference image are processed by a frozen text encoder and Variational autoencoder(VAE) encoder, respectively, yielding text tokens $c_t$ and image tokens $c_i$.
For the geometric constraints provided by the mesh, we render it as normal maps $n$, serving as geometric conditions. These are encoded into geometric features through a geometry encoder $G$ and injected into the multi-view noisy latents $x \in \mathbb{R}^{b \times f \times c \times h \times w}$, where $b$ and $f$ represent batch axis and number of views axis respectively, formatted as
\begin{equation}
    x_g = G(n) + x.
\end{equation}
After obtaining the noisy latents with geometric control signals, $x_g \in \mathbb{R}^{b \times f \times c \times h \times w}$, we integrate the image and text reference information with the geometric guidance. Specifically, we first replicate the reference image features according to the number of views and then reshape the view dimension of multi-view noisy latents $x_g$ to the batch dimension. We concatenate the view features with the reference information and apply appearance self-attention to fuse the reference information into each view based on the geometric guidance, resulting in the fused noisy latents $x_f \in \mathbb{R}^{(b \times f) \times c \times h \times w}$. Mathematically, 
\begin{equation}
\begin{aligned}
    Q^\text{app} &= W^\text{app}_{Q}(x_f), \\
    K^\text{app} &= W^\text{app}_{K}(x_f \oplus c_i \oplus c_t), \\
    V^\text{app} &= W^\text{app}_{V}(x_f \oplus c_i \oplus c_t),
\end{aligned}
\end{equation}
where $\oplus$ represents the concat operation. To further enhance the consistency of multi-view results, inspired by recent video generation work~\cite{yang2024cogvideoxtexttovideodiffusionmodels, liu2024sorareviewbackgroundtechnology}, we reshape $x_f$ back to $x_f \in \mathbb{R}^{b \times f \times c \times h \times w}$ to introduce a cross-frame global 3D attention mechanism to facilitate feature interaction and fusion across different views, improving spatial consistency in the generated outputs. Additionally, text and image reference information are included as feature tokens in the input, ensuring the continued preservation of reference appearance and semantics.

\paragraph{Multi-branch DiT for Material Generation.} As shown in Fig. \ref{fig:block}, we utilized the PBR metallic workflow in texture generation. This workflow defines materials using four maps: albedo, roughness, metallic, and normal. Given a specific mesh,
We introduce a multi-branch DiT to generate albedo, roughness, and metallic material properties under geometry conditions provided by the normal of the mesh.
In detail, we begin by inputting three sets of noisy latents, representing albedo, roughness, and metallic. After incorporating the normal map as a condition, we obtain $x_g^a$, $x_g^r$, and $x_g^m$ as multi-view noisy latents.
Then, We replicate the first three blocks of DiT to create three separate material branches: the albedo branch $B_a$, roughness branch $B_r$, and metallic branch $B_m$. Each set of latents, representing different material properties, is first projected through independent layers and then directed into its corresponding input branch. We obtain a shared latent $x_s$ by fusing the outputs of different material branches:

\begin{equation}
    x_s = \sum(B_a(x_g^a), B_r(x_g^r), B_m(x_g^m)),
\end{equation}

Afterward, the shared latent passes through a set of shared DiT blocks. These shared blocks enforce spatial and geometric consistency across different materials, ensuring unified appearance characteristics for each material corresponding to the mesh.
In the final three blocks of the DiT structure, similar to the input stage, we replicate the blocks to create three output branches. These branches take $x_s$ as input and decode the different materials to generate multi-view material results with 3D consistency.

\subsection{PBR-based Diffusion Loss}
\label{sec:pbrloss}
As proposed in ~\cite{salimans2022progressivedistillationfastsampling}, we employ V-Prediction and V-Loss during training, formatted as follows: 
\begin{equation}
    v_t = \sqrt{\overline{\alpha}_t}\epsilon - \sqrt{1 - \overline{\alpha}_t}x_0,
\end{equation}
\begin{equation}
    \mathcal{L}_v = \Vert v_t - \overline{v}_t \Vert,
\end{equation}
where $\epsilon$ is a Gaussian noise, $x_0$ is the original data, $\alpha$ represents a component of the variance schedule used to control noise in DDPM~\cite{ho2020denoising}, $\overline{\alpha}_t$ is the cumulative product of $\alpha_t$ from the initial time step up to time t, and $v_t$ is the velocity of $x_t = \sqrt{\overline{\alpha}_t}x_0 + \sqrt{1 - \overline{\alpha}_t}\epsilon$. 
Meanwhile, we note that albedo, roughness, and metallic each play distinct roles in the physically-based rendering (PBR). Therefore, in addition to the V-Loss, we introduce an additional PBR-based loss function, ensuring that the generated materials adhere to realistic physical principles. Specifically, after the model predicts $\overline{v}_t$ at a certain time step $t$, we can then compute the corresponding predicted $\overline{x}_0^t$ as:
\begin{equation}
    \overline{x}_0^t = \sqrt{\overline{\alpha}_t}x_t - \sqrt{1 - \overline{\alpha}_t}\overline{v}_t.
\end{equation}
Using the latent decoder $\mathcal{D}$,  we process each material branch to obtain the predicted original data at timestep $t$ for albedo $\mathbf{a}$, roughness $\mathbf{r}$, and metallic $\mathbf{m}$:
\begin{equation}
    \overline{\mathbf{a}}_0^t = \mathcal{D}(\overline{x_\mathbf{a}}_0^t),
    \overline{\mathbf{r}}_0^t = \mathcal{D}(\overline{x_\mathbf{r}}_0^t),
    \overline{\mathbf{m}}_0^t = \mathcal{D}(\overline{x_\mathbf{m}}_0^t).
\end{equation}
We set up a random lighting configuration and use a Cook-Torrance~\cite{10.1145/357290.357293} PBR renderer to render with the predicted $\mathbf{a}$, $\mathbf{r}$, $\mathbf{m}$, obtaining a PBR rendering output $\overline{x_\text{PBR}}_0$. By incorporating random lighting during training, we reduce the risk of shadows and lighting effects being embedded in the albedo, thus enhancing the accuracy of material generation. The PBR loss is then computed directly between this output and the ground-truth results rendered under the same lighting conditions, formatted as:
\begin{equation}
    \mathcal{L}_\text{PBR} = \Vert \overline{x_\text{PBR}}_0 - {x_\text{PBR}}_0 \Vert.
\end{equation}
Ultimately, our PBR-based diffusion loss consists of the V-Loss and the PBR loss, formatted as:
\begin{equation}
    \mathcal{L} = \mathcal{L}_v + \mathcal{L}_\text{PBR},
\end{equation}
The PBR-based diffusion loss function improves the accuracy of each texture and material branch while aligning with realistic, physically-based rendering principles.

\subsection{Material Refinement DiT}
\label{sec:uvdit}
After obtaining the multi-view results based on the reference image, text, and normal geometric constraints, we first use Blender~\cite{blender} smart-UV projection to unwrap the given mesh $\mathcal{M}$. Then, we un-project the generated materials onto the mesh, resulting in a coarse texture map $T_c$ based on the multi-view outputs. This coarse texture map has two main issues. First, the multi-view results do not fully cover the mesh, leading to some empty areas in the coarse texture map. Second, the relatively low-resolution multi-view images lack the necessary texture details for a high-resolution texture map, resulting in a loss of fine details.

To address these issues, We propose MR-DiT to further refine the texture. This DiT takes the coarse texture map $T_c$ as input and continues to optimize details in UV space. Inspired by tiled ControlNet~\cite{control-net}, we use the coarse texture map as a low-resolution tiled condition image, injecting it into the DiT by adding it to the noisy latents. Additionally, based on the smart-UV unwrapping results, we provide the normal map $T_n$ of the unwrapped mesh as an additional control condition. This map helps control the texture map generation region and serves as a geometric cue to guide the generation of texture details. The blocks in MR-DiT share a similar structure with those in MG-DiT, but with the Appearance Attention module removed, as control signals are directly incorporated. The refined texture map $T_r$ is obtained as:
\begin{equation}
    T_r = \mathcal{R}(z, c, T_c, T_n),
\end{equation}
where z is the noisy latents of MG-DiT and $c$ is the text corresponding to the mesh.




\begin{table*}[t]
\caption{Quantitative Comparisons: FID and KID are calculated with rendered GT; User study: numerical scores in four aspects assigned between 1 and 7 (higher is better).}

\centering
\begin{tabular}{ccccccccc}
\toprule
\multicolumn{2}{c}{Method} &TEXTure &Text2Tex&Paint3D&Fantasia3D&Paint-it &FlashTex&Ours \\
\midrule
\multicolumn{2}{c}{FID(↓)}                    & 44.86 & 61.39   & 33.62 & 77.04 & 37.22 & 39.90 & \textbf{31.97} \\
\multicolumn{2}{c}{KID(↓)($\times 10^{-3}$)}  & 6.45  & 5.93    & 4.95  & 8.57  & 4.76  & 4.29  & \textbf{4.16}  \\
\multirow{3}{*}{\begin{tabular}[c]{@{}c@{}}User \\ Study\end{tabular}}
& \multicolumn{1}{|c}{realism(↑)}  & 5.01 & 4.89 & 4.63 & 1.63 & 3.67 & 2.05  &\textbf{6.12} \\
& \multicolumn{1}{|c}{fidelity(↑)} & 5.12 & 4.90  &  4.16 & 2.56 & 3.78 & 1.54   & \textbf{5.94} \\
& \multicolumn{1}{|c}{generalization(↑)} & 2.38 & 2.54 & 3.09 & 4.22 & 5.00 & 4.44 & \textbf{6.33} \\
\bottomrule 
\end{tabular}
\label{table:Quantitative}
\end{table*}

\begin{figure*}[th]
    \centering
    \includegraphics[width=1.0\linewidth]{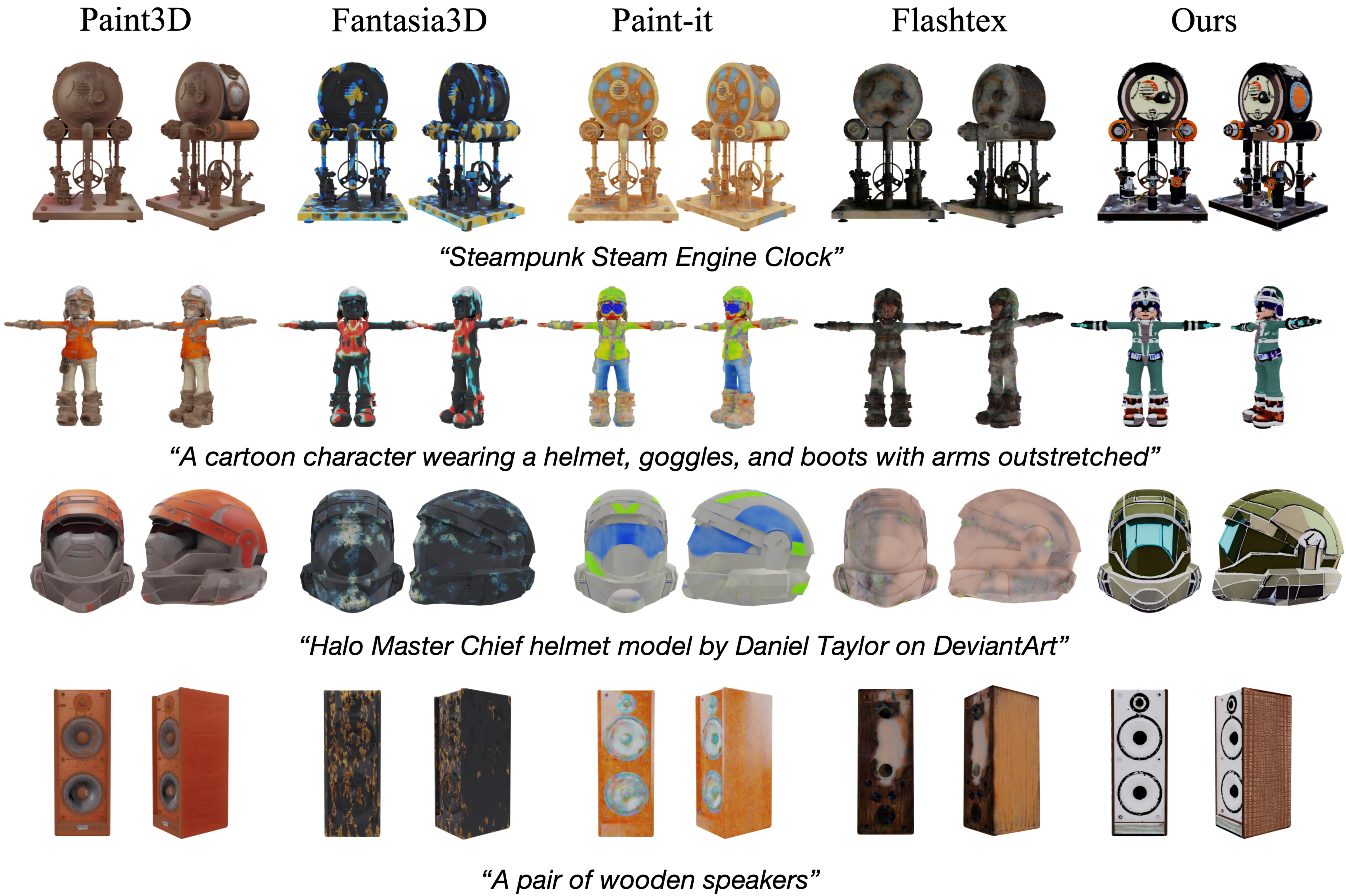}
    \caption{Qualitative comparisons on PBR material generation conditioned on text prompt.}
    \label{fig:quality_rets}
    \vspace{-5mm}
\end{figure*}

\begin{figure*}[th]
    \centering
    \includegraphics[width=1.0\linewidth]{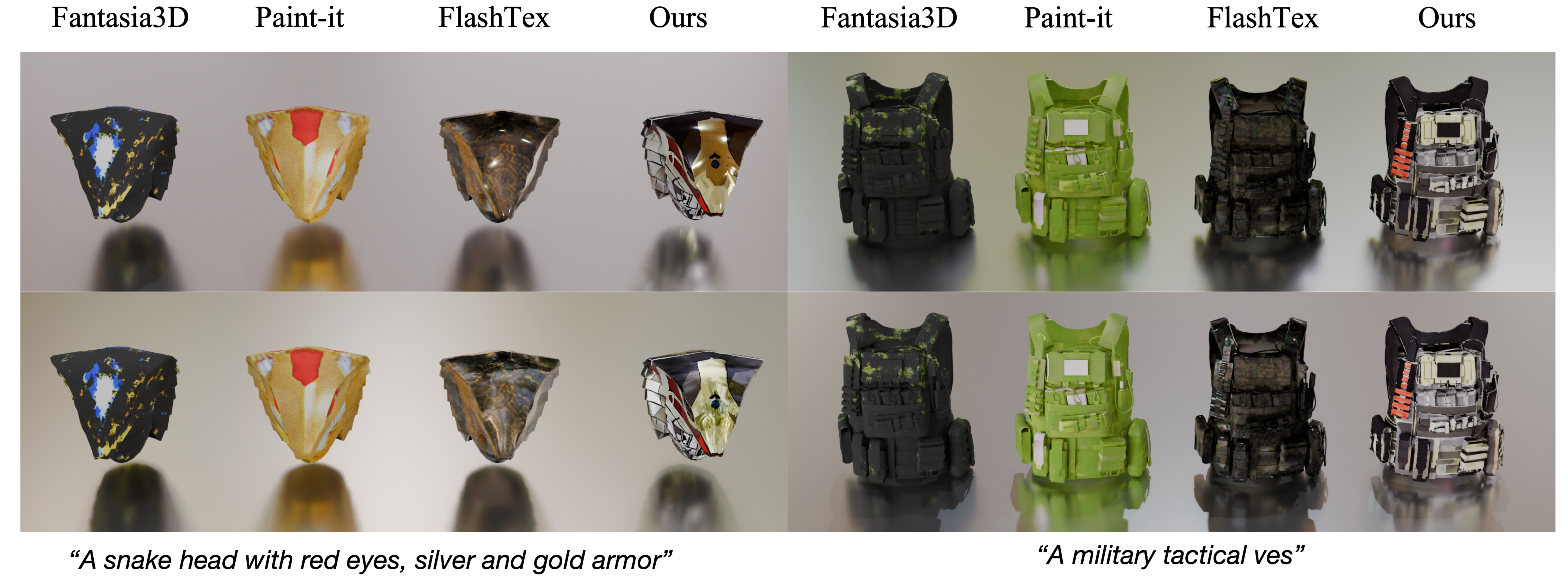}
    \caption{Qualitative comparisons on Relighting results}
    \label{fig:pbr_rets}
\end{figure*}

\section{Experiments}
We begin with the experimental setup in Sec. \ref{sec:exp_setup}. We then present the experimental results in Sec. \ref{sec:exp_rets}. Finally, we describe an ablation study in Sec. \ref{sec:ablation}.

\subsection{Experimental Setup}
\label{sec:exp_setup}

\noindent\textbf{Dataset.} We select 83,192 models across various categories from G-Objaverse~\cite{zuo2024sparse3d, qiu2024richdreamer, Objaverse} to train our MG-DiT. Normals from G-Objaverse serve as geometric control conditions, and we render ground truth values for albedo, roughness, and metallic properties in Blender. During texture refinement, we perform batch smart UV unwrapping, baking albedo, roughness, and metallic into ground truth PBR materials in UV space, along with normal-based geometric conditions. Lastly, we randomly project 12 viewpoints onto unwrapped UVs to obtain coarse UV map controls.

\noindent\textbf{Baseline Approaches.} We conduct a comparison of our approach against established mesh texturing methods. 
TEXTure~\cite{TEXTure_10.1145/3588432.3591503} creates 2D views from selected perspectives, which are conditioned on depth rendered from mesh, and then incrementally merge them into one texture by inversing projection. Text2Tex~\cite{chen2023text2tex} takes a similar approach but enhances the inpainting process and adds a texture refine stage. Paint3D~\cite{zeng2023paint3d} takes a coarse-to-fine generation approach: after the initial generation, it refines the textures using specialized diffusion models in UV space, removing lighting effects and enhancing texture quality. 
Moreover, methodologies for PBR texture generation are included as baseline comparisons. 
Fantasia3D~\cite{Chen_2023_ICCV_Fantasia3D} is chosen as one of the representative baselines, being the first to adopt a material-based representation for text-to-3D texturing. 
Paint-it~\cite{youwang2023paintit} uses synthesis-through-optimization, refining texture quality with deep convolutional re-parameterization, achieving high fidelity and photorealism through effective noise filtering. 
FlashTex~\cite{deng2024flashtex} employs a rendering engine to generate illumination-informed maps for training a LightControlNet model, which disentangles lighting from material properties for faster, higher-quality texture generation.

\noindent\textbf{Implementation Details.} For MG-DiT, our DiT model consists of 42 transformer blocks, including 24 shared blocks and 18 branch-specific blocks. We finetune the model based on CogvideoX-2B~\cite{yang2024cogvideoxtexttovideodiffusionmodels}. Our goal is to generate 512x512 videos with a length of 12 frames. We train the model with a learning rate of 2e-4 and a batch size of 1 on 16 A100 GPUs for 100,000 steps, taking approximately five days. In the refinement stage, our DiT model consists of 30 transformer blocks. We generate 2048x2048 resolution material maps, training the model with a learning rate of 2e-4 and a batch size of 1 on 16 A100 GPUs for 50,000 steps, which takes approximately two days. 

\subsection{Experimental Results}
\label{sec:exp_rets}
We conducted qualitative and quantitative comparisons on text-to-texture tasks and a user study. Experimental results demonstrate that our method can generate lightless yet high-quality 2K PBR materials based on textual descriptions.

\noindent\textbf{Qualitative Comparisons on Text-to-material.}
As shown in Fig. \ref{fig:quality_rets}, in qualitative experiments, we compare our method with PBR material generation approaches, including Fantasia3D~\cite{Chen_2023_ICCV_Fantasia3D}, Paint-it~\cite{youwang2023paintit}, and Flashtex~\cite{deng2024flashtex}, rendering the generated PBR material under identical environment lighting. Additionally, we conduct comparisons with a texture-only text-to-texture generation method Paint3D~\cite{zeng2023paint3d}. We use normal ControlNet~\cite{control-net} based on text input in the experiments to generate the reference image. We used a uniformly distributed ambient lighting setup to render the results for a fair comparison. Experimental results show that texture-only generation methods often embed ambient lighting effects into the results, leading to poor adaptability under new lighting conditions. In comparison, our approach leverages the MG-DiT model to address inconsistencies, reduce artifacts, and improve clarity. By integrating textual and image references, our method further produces PBR materials with detailed textures and vibrant colors. 

\noindent\textbf{Qualitative Comparisons on Relighting.}
As illustrated in Fig. \ref{fig:pbr_rets}, we compared the relighting effects under different ambient lighting conditions between our method and baseline approaches for PBR material generation. The experimental results indicate that our method achieves the closest alignment with textual descriptions, producing textures of higher quality than baseline methods. Additionally, it demonstrates strong generalization across scenes with varying ambient lighting conditions.

\noindent\textbf{Quantitative Comparisons.}
Quantitative comparisons of our method against baseline approaches are conducted on a test set comprising 70 models that are randomly selected from G-Objaverse~\cite{zuo2024sparse3d,qiu2024richdreamer,Objaverse}.
Following FlashTex~\cite{deng2024flashtex}, we assess the quality using Frechet Inception Distance (FID) and Kernel Inception Distance (KID) relative to ground-truth renderings. For each model, 16 views are generated. As demonstrated in Table. \ref{table:Quantitative}, the numerical results demonstrate that our method outperforms the baseline approaches in PBR material generation quality and accuracy.

\noindent\textbf{User Study.} We conduct a user study to evaluate the realism of the rendered textures, their fidelity to the input text prompts, and their generalization performance under varying lighting conditions. We randomly selected 20 meshes and their corresponding text prompts for this study. Each mesh was textured by both Paint3D and the baseline models and presented to users randomly. Each textured object was shown in full detail through a 360-degree rotation view. Participants were asked to rate the results on three aspects: (1) realism of the rendered textures, (2) fidelity to the text prompt, and (3) generalization performance under different lighting conditions, using a 1-to-7 scale. We gathered responses from 30 users, with the average ratings for each method shown in Table. ~\ref{table:Quantitative}. As demonstrated, our method outperforms the baselines in all three aspects.

\begin{figure}
    \includegraphics[width=0.98\linewidth]{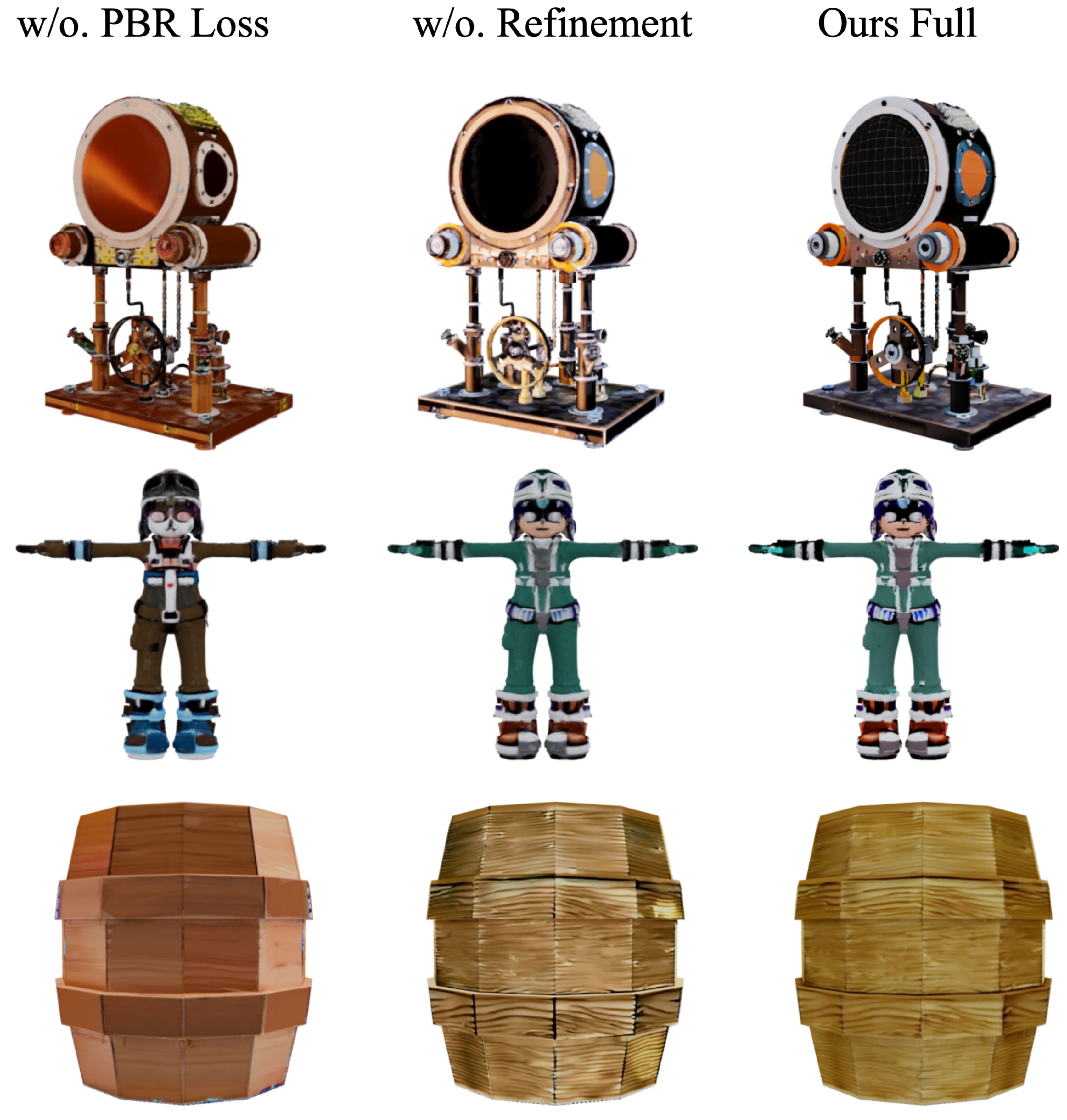}
    \caption{Ablation study. The proposed PBR-based Diffusion Loss and Material Refinement DiT enhance the visual qualities and material accuracy of the meshes.}
    \label{fig:ablation}
\end{figure}

\subsection{Ablation Study}
\label{sec:ablation}

As shown in Fig. \ref{fig:ablation} and Table. \ref{tab:ablation}, We conduct both quantitative and qualitative ablation studies based on the following settings:

(a) Without PBR-Based Diffusion Loss: When the PBR-based diffusion loss is omitted, the model overfits the lighting and shading effects in the reference images. This results in mixed lighting information on the albedo map and reduces the consistency in the generated roughness and metallic maps.

(b) Without MR-DiT: Without the Material Refinement DiT, the resolution of the generated textures is insufficient, leading to noticeable detail loss. Additionally, due to the limited number of views in multi-view generation, regions not covered by the viewpoints exhibit significant artifacts.

\begin{table}
\caption{Ablation study}
\small
\centering
\begin{tabular}{cccc}
\toprule
Ablation  & w/o. PBR Loss & w/o. MR-DiT & Full Model 
\\
\midrule
FID(↓) &37.22&34.55&31.97 \\
KID(↓) &4.84&4.32&4.16 \\ 
\bottomrule
\end{tabular}
\label{tab:ablation}
\end{table}

\section{Conclusion}
\label{sec:onclusion}

In this work, we propose a novel approach for generating multi-view-consistent and physically accurate PBR materials for 3D models. Our method introduces a two-stage pipeline, utilizing a Multi-View Generation Diffusion Transformer (MG-DiT) to ensure spatial consistency across different views and employing a Material Refinement DiT (MR-DiT) to enhance detail and completeness in the UV space. Through comprehensive experiments, we demonstrate that our approach achieves state-of-the-art performance in generating high-resolution PBR materials that are visually appealing and adaptable to various lighting conditions, making it suitable for realistic rendering applications.

\paragraph{Limitations and Future Work.}
Our method has two main limitations. The two-stage pipeline requires significant computational resources, which makes it difficult to apply in real-time or large-scale scenarios. Additionally, the inconsistent representation of material predictions, especially for text-based inputs, can reduce accuracy. Future work will focus on designing more efficient architectures to reduce computational demands. It will also aim to create unified representations for text-based inputs to improve prediction accuracy and practical usability.
{
    \small
    \bibliographystyle{ieeenat_fullname}
    \bibliography{main}
}

\maketitlesupplementary

\appendix


In the supplementary material, we first provide a detailed explanation of the implementation and relighting aspects of our method (Sec. \ref{sec:supp_detail}). Next, we present additional experimental results that were not included in the main text due to space constraints (Sec. \ref{sec:supp_additional}). Finally, we discuss the societal impact of our method (Sec. \ref{sec:supp_social}).

\section{Details of the Method}
\label{sec:supp_detail}

\subsection{Implementation Details}
\paragraph{Base Model} We employ CogVideoX-2B~\cite{yang2024cogvideoxtexttovideodiffusionmodels} For the Multi-View Generation DiT (MG-DiT) and Material Refinement DiT (MR-DiT), which consists of 30 DiT blocks in total. In MG-DiT, the first three and the last three DiT blocks are duplicated three times for the three material branches to initialize the respective branches. The remaining 24 DiT blocks are used to initialize the shared DiT blocks in the middle. We use T5~\cite{raffel2023exploringlimitstransferlearning} as the text encoder to obtain the corresponding text tokens.

\paragraph{Training Details} We train the MG-DiT model at a resolution of 512$\times$512, resulting in a feature map size 64$\times$64 after VAE encoding. The hidden size is set to 1920, with 32 attention heads used in the attention mechanism. The training is conducted with the following hyperparameter settings: a time embedding size of 256, weight decay set to 1e-4, and Adam optimizer parameters configured as $\epsilon$=1e-8, $\beta$=0.9, and $\beta$=0.95. The learning rate follows a cosine decay schedule, with gradient clipping applied at 1.0. The text input length is fixed at 226, and the maximum sequence length is capped at 82k tokens. A minimum aesthetic value threshold of 4.5 is applied, and training is performed with FP16 precision. MR-DiT adopts a training strategy similar to MG-DiT, with the main difference being an increase in feature map size from 64$\times$64 to 256$\times$256. This adjustment enables the final output of materials at a resolution of 2048$\times$2048. We use the Zero-SNR~\cite{lin2024commondiffusionnoiseschedules} method to design a customizable noise schedule for DDPMs~\cite{ho2020denoising}. This approach constructs a linear schedule for noise variance and dynamically adjusts the signal-to-noise ratio (SNR) to control the diffusion process. It ensures a smooth progression across timesteps while maintaining flexibility for fine-tuning the model’s behavior during training and inference.

\paragraph{Inference Details} During the inference stage of MG-DiT and MR-DiT, the three types of materials are each initialized with Gaussian noise. We use the DPM++ solver~\cite{lu2023dpmsolverfastsolverguided} to sample in 50 steps to obtain the final results. We use Dynamic Classifier-Free Guidance (CFG) to adjust the guidance scale during sampling. The guidance scale follows a cosine-based schedule, transitioning smoothly as sampling progresses. This method is configured with a scale parameter of 6 and an exponential power of 5 to control the rate of change, providing adaptive guidance throughout the generation process. 

\paragraph{Loss Details} During the training of MG-DiT and MR-DiT, we utilize V-Loss as the standard diffusion loss. In the process of calculating the PBR-based diffusion loss for MG-DiT, we randomly initialize 3-10 point light sources in the spatial domain, with their intensities randomly chosen within the range of 1-10. Using the decoded material properties, we employ a differentiable PBR renderer to generate renderings under the specified lighting conditions. Similarly, the ground-truth materials are rendered under the same lighting setup. This process produces both the generated results and the ground-truth results under random lighting conditions.

\subsection{Relighting Details}
We use the Cook-Torrance model~\cite{10.1145/357290.357293} in our PBR-based diffusion loss; the Cook-Torrance model is a physically based rendering framework that combines microfacet theory with a bidirectional reflectance distribution function (BRDF) to simulate light interaction with surfaces accurately. The model decomposes the light reflection into diffuse and specular components. In our implementation, The rendering equation is expressed as:
\vspace{-0.3mm}
\begin{equation}
     L_o(\mathbf{p}, \omega_o) = \int_\Omega f_r(\mathbf{p}, \omega_i, \omega_o) L_i(\mathbf{p}, \omega_i) (\mathbf{n} \cdot \omega_i) \, \mathrm{d}\omega_i, 
\end{equation}

where \( L_o \) is the outgoing radiance at a surface point \( \mathbf{p} \) in the direction \( \omega_o \), and \( L_i \) represents the incoming radiance from the direction \( \omega_i \). The Bidirectional Reflectance Distribution Function(BRDF), \( f_r \), consists of a diffuse term and a specular term. The diffuse term follows Lambertian reflectance, expressed as \( f_{\text{diffuse}} = \frac{\mathbf{a}}{\pi} \), where \( \mathbf{a} \) is the albedo. The specular term uses the Cook-Torrance microfacet BRDF, defined as:

\begin{equation}
     f_{\text{specular}} = \frac{D(h) G(\omega_i, \omega_o) F(\omega_o)}{4 (\mathbf{n} \cdot \omega_i)(\mathbf{n} \cdot \omega_o)}.
\end{equation}

 In this equation, \( D(h) \) is the normal distribution function (NDF), modeled as:
 \begin{equation}
      D(h) = \frac{\alpha^2}{\pi ((\mathbf{n} \cdot h)^2 (\alpha^2 - 1) + 1)^2},
 \end{equation}
 where \( h \) is the normalized half-vector \( h = \frac{\omega_i + \omega_o}{\|\omega_i + \omega_o\|} \), and \( \alpha=\textbf{r}^2 \) represents the square of roughness. The geometry function, \( G(\omega_i, \omega_o) \), accounts for the shadowing and masking effects of microfacets, defined as:
 \begin{equation}
       G(\omega_i, \omega_o) = \min\left(1, \frac{2 (\mathbf{n} \cdot h) (\mathbf{n} \cdot \omega_o)}{\omega_o \cdot h}, \frac{2 (\mathbf{n} \cdot h) (\mathbf{n} \cdot \omega_i)}{\omega_o \cdot h}\right).
 \end{equation}
 
 Finally, the Fresnel term, \( F(\omega_o) \), captures the proportion of light reflected at the surface, expressed as:
 \begin{equation}
      F(\omega_o) = F_0 + (1 - F_0)(1 - (\mathbf{h} \cdot \omega_o))^5,
 \end{equation}
  where 
$F_0 = \mathbf{m} \cdot \mathbf{a} + (1 - \mathbf{m}) \cdot 0.04$
 is the reflectance at normal incidence, $\mathbf{m}$ is the metallic. This formulation allows the Cook-Torrance model to realistically render materials by accounting for the roughness $\mathbf{r}$, metallic $\mathbf{m}$, light directionality, and Fresnel effects. The final BRDF $f_r$ used in the rendering equation is expressed as:
\begin{equation}
    f_r(\mathbf{p}, \omega_i, \omega_o) = (1 - \mathbf{m})f_{\text{diffuse}} + f_{\text{specular}}
\end{equation}

\begin{figure*}[th]
    \centering
    \includegraphics[width=0.95\linewidth]{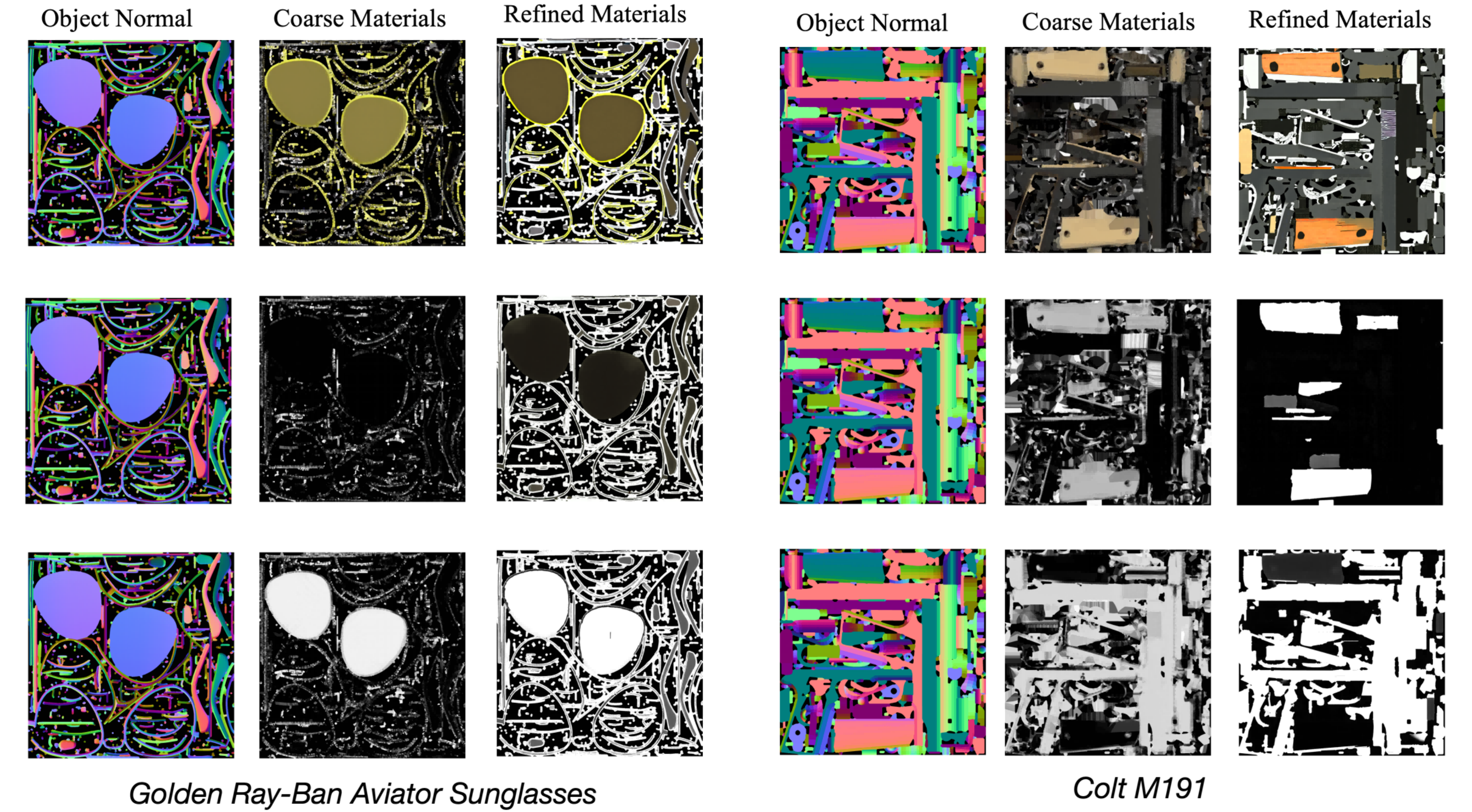}
    \caption{Visual results of Material Refinement: Albedo (top), Roughness (middle), and Metallic (bottom)}
    \label{fig:supp_refine}
\end{figure*}

\begin{figure*}[th]
    \centering
    \includegraphics[width=0.9\linewidth]{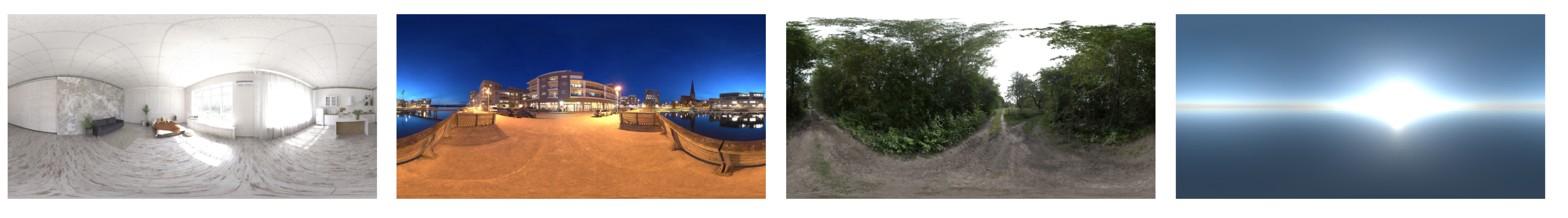}
    \caption{Environment maps for relighting}
    \label{fig:supp_lights}
\end{figure*}

\vspace{-0.2mm}
\section{Additional Results}
\label{sec:supp_additional}

\subsection{Multi-View Material Generation}

\begin{figure*}[th]
    \centering
    \includegraphics[width=0.95\linewidth]{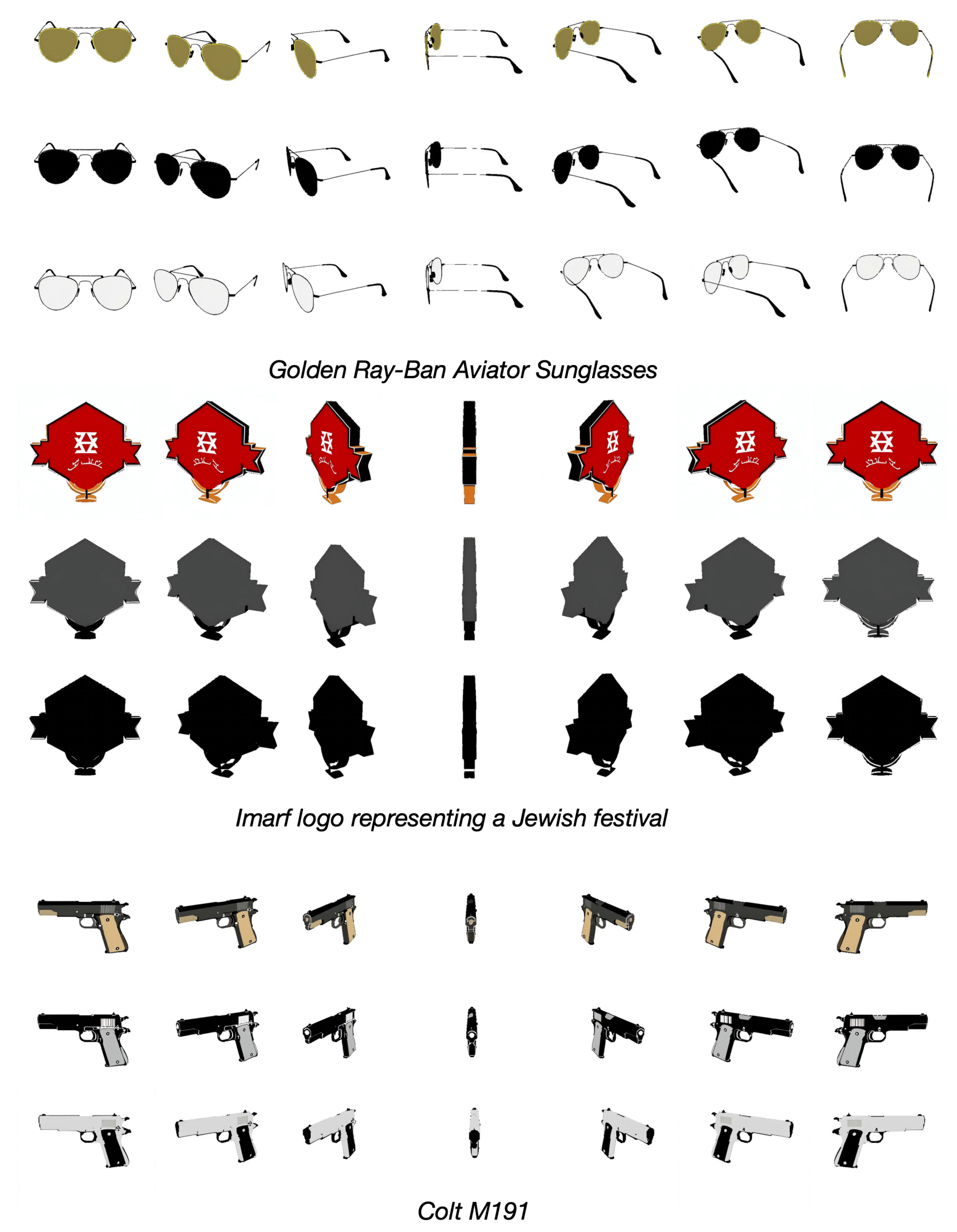}
    \caption{Visual results of Multi-View material generation: Albedo (top), Roughness (middle), and Metallic (bottom) (Part 1)}
    \label{fig:supp_mv_0}
\end{figure*}

\begin{figure*}[th]
    \centering
    \includegraphics[width=0.95\linewidth]{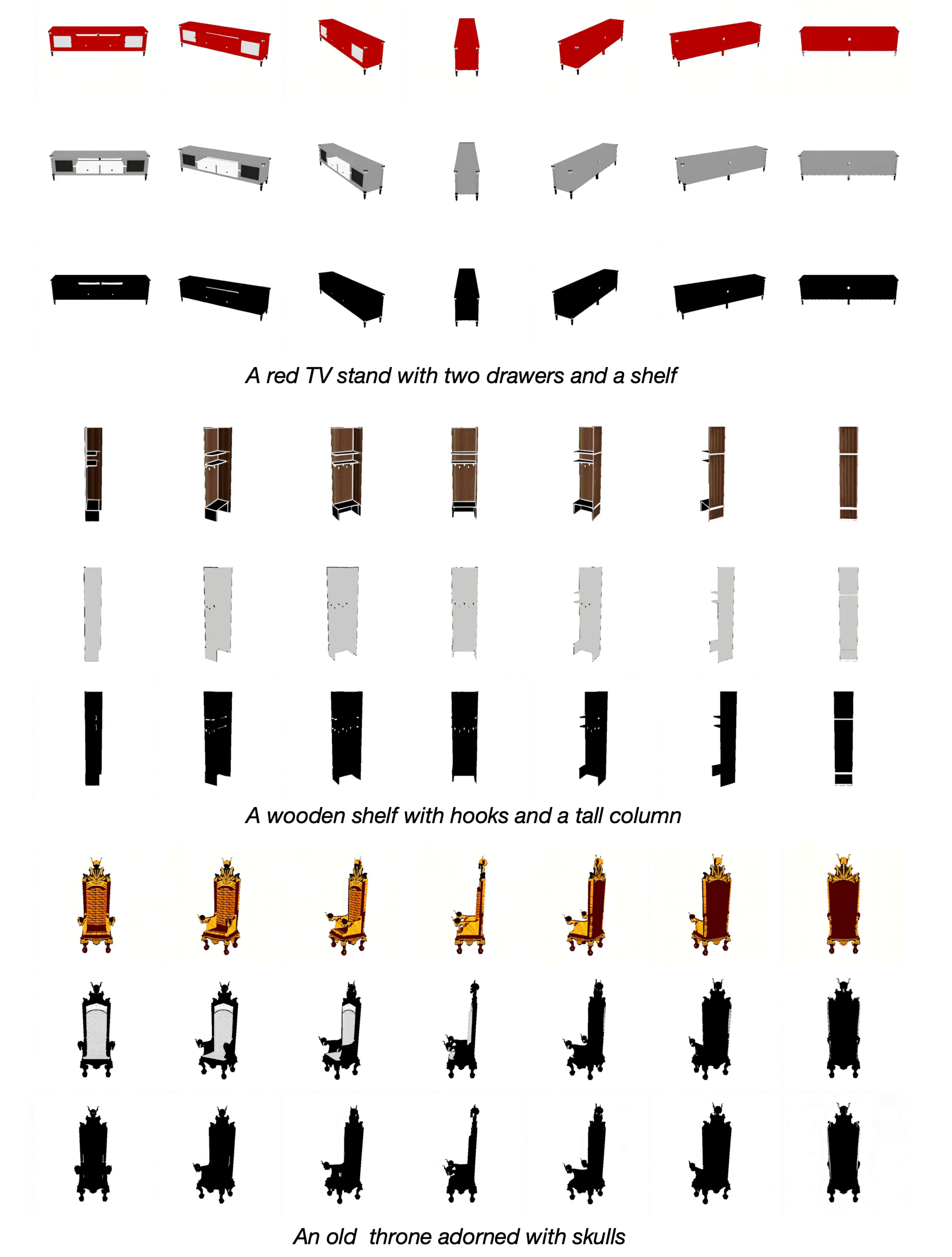}
    \caption{Visual results of Multi-View material generation: Albedo (top), Roughness (middle), and Metallic (bottom) (Part 2)}
    \label{fig:supp_mv_1}
\end{figure*}

In Fig. \ref{fig:supp_mv_0} and \ref{fig:supp_mv_1}, we present the results of Multi-View Material Generation. Our MG-DiT demonstrates the ability to generate light-independent albedo and accurate material properties while maintaining high multi-view consistency

\subsection{Material Refinement}

As shown in Fig. \ref{fig:supp_refine}, our MR-DiT refines the generated materials, enhancing their realism and clarity while completing mesh regions not covered during the Multi-View Generation process. This process ultimately produces high-quality materials with a resolution of 2K.

\subsection{More Textured Meshes}
\begin{figure*}[th]
    \centering
    \includegraphics[width=0.95\linewidth]{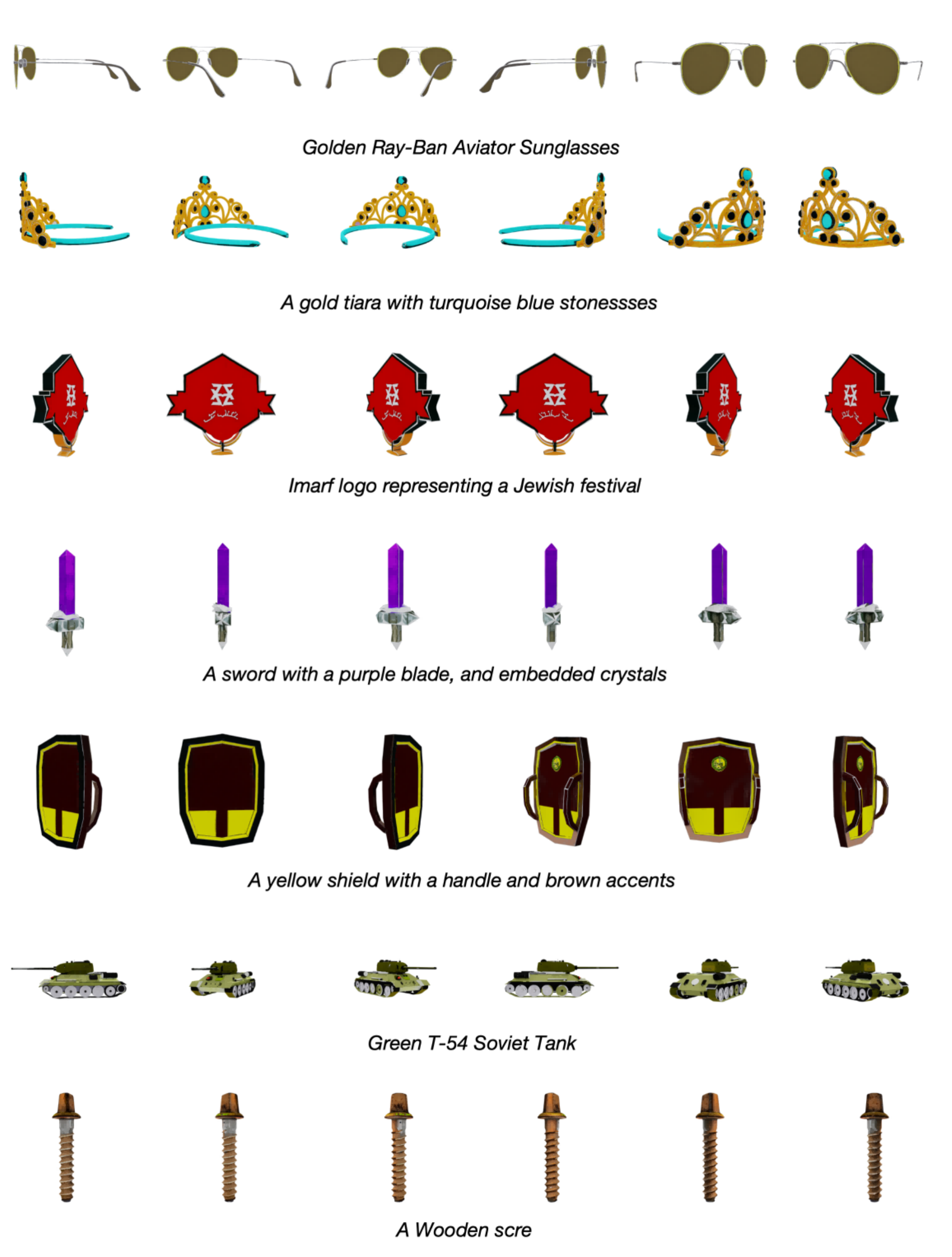}
    \caption{Visual results of textured meshes using generated materials (Part 1)}
    \label{fig:supp_tex_0}
\end{figure*}

\begin{figure*}[th]
    \centering
    \includegraphics[width=0.88\linewidth]{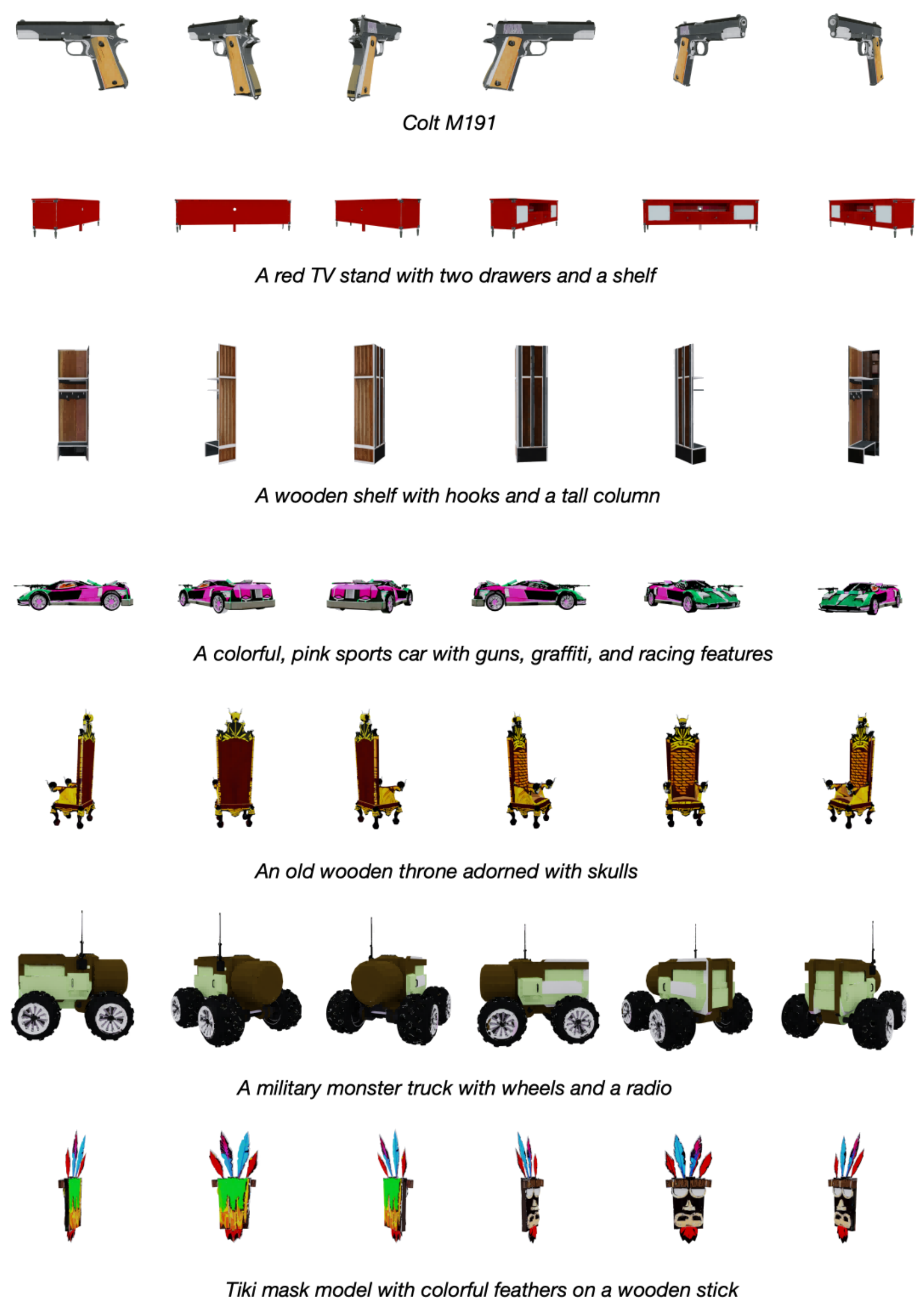}
    \caption{Visual results of textured meshes using generated materials (Part 2)}
    \label{fig:supp_tex_1}
\end{figure*}

We present additional textured meshes using the generated material results in Fig. \ref{fig:supp_tex_0} and \ref{fig:supp_tex_1}. The visualization demonstrates that our method can produce high-quality materials with strong generalization capability across various types of 3D models.

\subsection{More Relighting Results}
\begin{figure*}[th]
    \centering
    \includegraphics[width=0.88\linewidth]{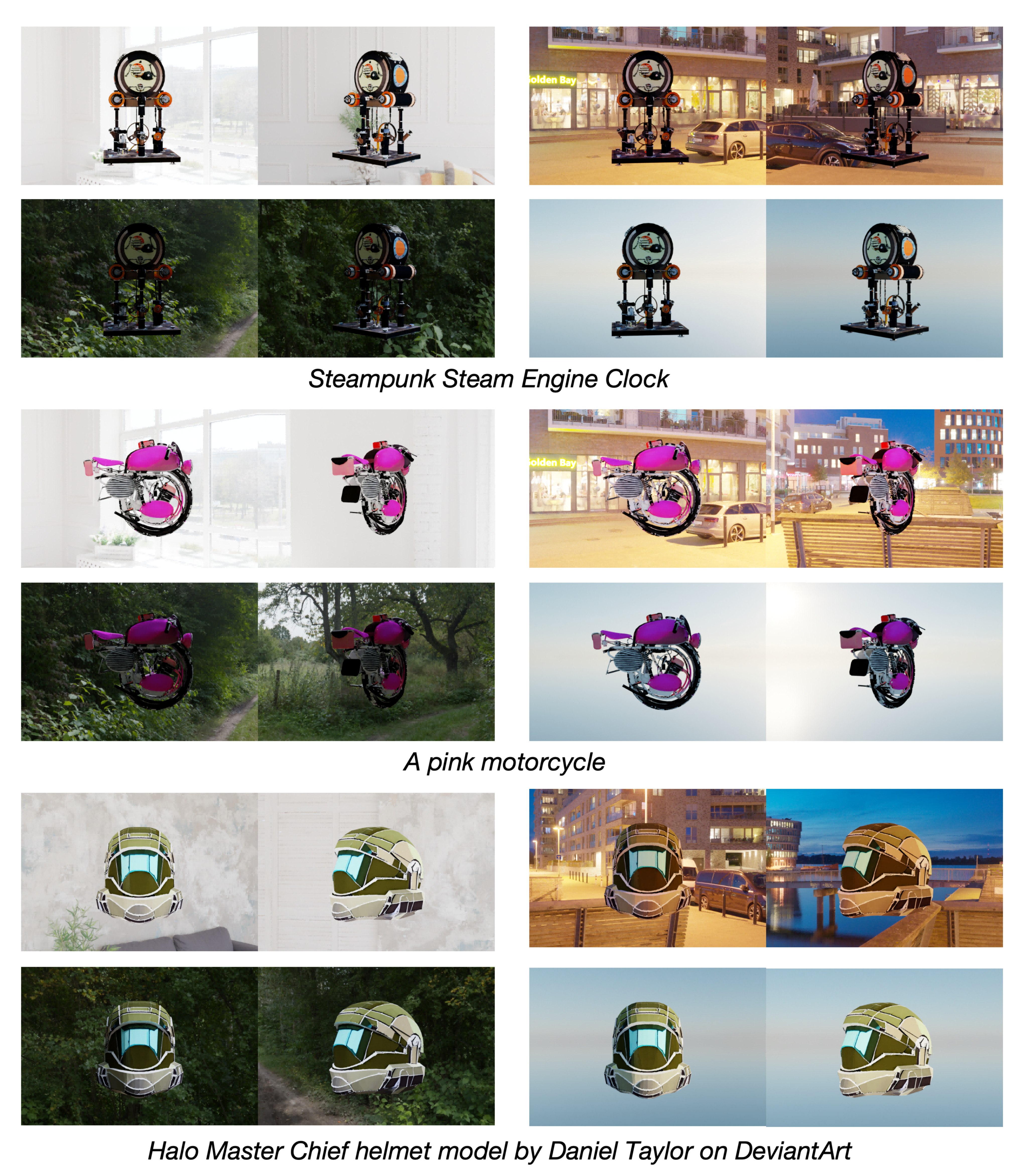}
    \caption{Visual results of relighting}
    \label{fig:supp_relighting}
\end{figure*}

As shown in Fig. \ref{fig:supp_relighting}, we apply relighting to the meshes textured with our generated materials under different environment lighting conditions in Fig. \ref{fig:supp_lights}. The relighting results further demonstrate the accuracy of our material generation while reducing artifacts attributed to the high multi-view consistency. Additional visual results are presented in the supplementary video.

\section{Societal Impact}
\label{sec:supp_social}
Our work in material generation represents a significant leap forward in automating and enhancing the material creation process for 3D content, making it faster, more accessible, and more efficient. This technology can potentially revolutionize industries such as gaming, virtual reality, architecture, and digital entertainment by lowering technical barriers and reducing the time and computational costs associated with traditional material authoring. Furthermore, the improved efficiency of our method supports more sustainable practices by minimizing energy consumption and resource waste. However, alongside these benefits, there are notable risks and challenges. One concern is the potential misuse of this technology to create highly realistic but deceptive or malicious content, such as fake objects, environments, or even fraudulent products, which could contribute to misinformation or unethical practices. The ability to generate hyper-realistic materials could also raise intellectual property concerns, as it may enable the replication of copyrighted designs without authorization. Additionally, democratizing advanced material generation tools could disrupt traditional industries and reduce demand for skilled artisans, leading to economic and employment challenges in certain sectors. While our method currently requires careful curation and validation to ensure high-quality results, these concerns highlight the need for responsible usage and the development of safeguards to prevent misuse.

\end{document}